\documentclass[conference]{IEEEtran}
\IEEEoverridecommandlockouts
\usepackage{cite}
\usepackage{amsmath,amssymb,amsfonts}
\usepackage{algorithmic}
\usepackage{graphicx}
\usepackage{textcomp}
\usepackage{xcolor}
\usepackage{times}
\def\BibTeX{{\rm B\kern-.05em{\sc i\kern-.025em b}\kern-.08em
    T\kern-.1667em\lower.7ex\hbox{E}\kern-.125emX}}

\makeatletter
\renewcommand\IEEEkeywordsname{Keywords}
\makeatother
\usepackage{xpatch}
\xpatchcmd\IEEEkeywords{---}{-}{}{}

\usepackage{caption}
\usepackage[switch]{lineno}

\makeatletter
\renewcommand{\fnum@figure}{Figure~\thefigure}
\makeatother

\usepackage{graphicx}
\usepackage[hidelinks]{hyperref}
\usepackage{color}

\usepackage{makecell}

\usepackage{calrsfs}
\DeclareMathAlphabet{\pazocal}{OMS}{zplm}{m}{n}

\usepackage{booktabs}
\usepackage{multirow}
\usepackage{tabularx}
\usepackage{amssymb,amsfonts,amsmath}
\DeclareMathOperator*{\argmax}{arg\,max}

\usepackage{cite}

\begin{document}

\title{{\bfseries\Large Camera Model Identification Using Audio and Visual Content from Videos}\\

\author{\IEEEauthorblockN{Ioannis Tsingalis, Christos Korgialas, Constantine Kotropoulos}
\IEEEauthorblockA{Department of Informatics Aristotle University of Thessaloniki\\
Thessaloniki 54124, Greece \\
Email: {tsingalis, ckorgial, costas}@csd.auth.gr}
}}

\maketitle

\begin{abstract}

The identification of device brands and models plays a pivotal role in the realm of multimedia forensic applications. This paper presents a framework capable of identifying devices using audio, visual content, or a fusion of them. The fusion of visual and audio content occurs later by applying two fundamental fusion rules: the product and the sum. The device identification problem is tackled as a classification one by leveraging Convolutional Neural Networks. Experimental evaluation illustrates that the proposed framework exhibits promising classification performance when independently using audio or visual content. Furthermore, although the fusion results don't consistently surpass both individual modalities, they demonstrate promising potential for enhancing classification performance. Future research could refine the fusion process to improve classification performance in both modalities consistently. Finally, a statistical significance test is performed for a more in-depth study of the classification results. 
\end{abstract}

\begin{IEEEkeywords}
\textbf{\textit{Camera Model Identification (CMI); Convolutional Neural Networks (CNNs); Sum and Product Fusion Rules; Statistical Testing; Multimedia Forensics.}}
\end{IEEEkeywords}

\section{Introduction}

Camera Model Identification (CMI) \cite{berdich2023survey}\cite{nwokeji2024source} emerges as an essential forensic tool, particularly in the pursuit of discerning the brand or model of a mobile phone from a recording \cite{stamm2013information} \cite{diwan2023visualizing}. The forensic analysis delves into various multimedia elements, including audio recordings, images, and videos, to unravel the distinct characteristics and signatures of different mobile phone brands/models. 
By exploiting these signatures, forensic analysts can accurately determine the particular device that recorded the multimedia content, providing crucial insights into various investigations, such as identifying the perpetrators behind a felony scene.

Two prominent types of signatures employed in device identification are Photo-Response Non-Uniformity (PRNU) \cite{lukas2006digital} for images and Mel-Frequency Cepstral Coefficients (MFCCs) \cite{davis1980comparison}\cite{kotropoulos2014source}\cite{kotropoulos2014mobile}\cite{Kritsiolis2024} extracted from audio recordings. PRNU analysis involves studying the unique noise patterns present in images, allowing forensic experts to identify the camera model with high precision. On the other hand, MFCCs, extracted from the audio recorded by a mobile phone speaker, serve as distinctive ``fingerprints" that enable analysts to discern which mobile device is used for recording. 

Both methodologies contribute substantially to the forensic toolkit, offering valuable and intricate details regarding multimedia content's recording time and place. This encompasses insights into its creation process, source, authenticity, and other pertinent characteristics.

However, the evolution of deep learning has catalyzed a notable shift in research focus, particularly emphasizing the application of Convolutional Neural Networks (CNNs) in extracting inherent patterns from multimedia content \cite{bondi2016first}. 

This advancement has significantly enhanced the ability to classify and identify devices by analyzing raw video frames and log-Mel spectrograms as key inputs of CNNs, as described in Section~\ref{subse:featureExtraction}. 

Consequently, this approach has expanded the scope of modalities used, going beyond traditional PRNU and MFCC analysis to incorporate a broader spectrum of features. The integration of CNNs marks a pivotal stride in the ongoing refinement of forensic techniques, offering a framework for device identification. The framework combines conditional probability densities of device identification given the audio and visual content in a late fusion manner, hoping to overcome any caveats when one of the two modalities is employed for CMI (i.e.,  a high noise regime in the visual content).

\noindent \textbf{Motivation and Contribution.} Inspired by the application of CMI in forensics, this paper introduces a framework for CMI, treating it as a classification problem. CNNs trained on either audio or visual content are employed for this purpose. Experimental findings showcase promising performance when employing either audio or visual content individually. Furthermore, late fusion integrates the decision given the audio and visual content by utilizing fundamental fusion rules, namely the product and sum rule~\cite{kittler1998}. 
Applying these rules for classification offers valuable insights for future research in the fusion of modalities for CMI. Given the limited existing research in this area, this work represents a significant contribution to the literature, paving the way for further exploration. The code for the proposed framework can be found at \cite{IARIADevIDFusion}.

The remaining paper is organized as follows. In Section~\ref{se:related}, a survey of related works is undertaken. In Section~\ref{se:dataset}, the dataset is described. Section~\ref{se:framework} outlines the proposed methodology with experimental results presented and discussed in Section~\ref{se:evaluation}. Finally, the paper is concluded in Section \ref{se:conclusion}, discussing the results obtained and outlining potential methods for future research.
\captionsetup{font={footnotesize,sc},justification=centering,labelsep=period}%
\begin{table*}[!ht]
    \caption{The 35 devices featured in the VISION dataset.}
    \label{tab:vision}
    \setlength{\tabcolsep}{20pt}
    \centering
    \begin{tabular}{clcl}
        \toprule
        \textbf{ID} & \textbf{Model} & \textbf{ID} & \textbf{Model} \\
        \midrule
        D01 & Samsung Galaxy S3 Mini & D19 & Apple iPhone 6 Plus \\
        D02 & Apple iPhone 4s & D20 & Apple iPad Mini \\
        D03 & Huawei P9 & D21 & Wiko Ridge 4G \\
        D04 & LG D2 90 & D22 & Samsung Galaxy Trend Plus \\
        D05 & Apple iPhone 5c & D23 & Asus Zenfone 2 Laser \\
        D06 & Apple iPhone 6 & D24 & Xiaomi Redmi Note 3 \\
        D07 & Lenovo P70 A & D25 & OnePlus A3000 \\
        D08 & Samsung Galaxy Tab 3 & D26 & Samsung Galaxy S3 \\
        D09 & Apple iPhone 4 & D27 & Samsung Galaxy S5 \\
        D10 & Apple iPhone 4s & D28 & Huawei P8 \\
        D11 & Samsung Galaxy S3 & D29 & Apple iPhone 5 \\
        D12 & Sony Xperia Z1 Compact & D30 & Huawei Honor 5c \\
        D13 & Apple iPad 2 & D31 & Samsung Galaxy S4 Mini \\
        D14 & Apple iPhone 5c & D32 & OnePlus A3003 \\
        D15 & Apple iPhone 6 & D33 & Huawei Ascend \\
        D16 & Huawei P9 Lite & D34 & Apple iPhone 5 \\
        D17 & Microsoft Lumia 640 LTE & D35 & Samsung Galaxy Tab A \\
        D18 & Apple iPhone 5c & & \\
        \bottomrule
    \end{tabular}
\end{table*}
\captionsetup{font={footnotesize,rm},justification=centering,labelsep=period}%

\section{Related Work}\label{se:related}

Research on brand device identification has focused on examining camera video sequences to ensure accurate recognition. In \cite{timmerman2020video}, an approach to CMI from videos was presented, utilizing extended constrained convolutional layers for extracting camera-specific noise patterns from color video frames. The approach offered robustness against compression techniques like WhatsApp and YouTube. An algorithm was proposed in \cite{lopez2017smartphone} for the CMI of the mobile device that created a video, utilizing sensor noise and wavelet transform for identification. The experiments demonstrated its effectiveness. In \cite{mandelli2020modified}, an algorithm addressing geometric misalignment in device brand identification was introduced, leveraging frequency domain searches for scaling and rotation parameters to efficiently align characteristic noise patterns with camera sensor traces, employing real videos from a benchmark dataset. 
Moreover, in \cite{altinisik2022video}, a CMI method was elaborated, incorporating encoding and encapsulation aspects into a joint metadata framework and employing a two-level hierarchical classification to achieve a  91\% accuracy in identifying video classes among over 20,000 videos from four public datasets. In \cite{akbari2022prnu}, a CNN named PRNU-Net, integrating a PRNU-based layer for source camera identification, was developed in response to the security challenges posed by the widespread distribution of digital videos, demonstrating competitive performance by emphasizing low-level features. Deep learning methods were applied to the identification of source camera devices from digital videos in \cite{bennabhaktula2022source}, achieving record accuracies on the VISION \cite{shullani2017vision} and QUFVD \cite{akbari2022new} datasets without the constraints of traditional PRNU-noise-based approaches. In \cite{manisha2023source}, an approach was introduced to address the challenges of video-based source camera identification, exacerbated by compression artifacts and pixel misalignment, by leveraging a resilient global stochastic fingerprint in the low- and mid-frequency bands.

Additionally, fusion techniques were developed, employing multiple modalities further to enhance the robustness and accuracy of CMI tasks. In \cite{hosler2019video}, a deep learning-based system was introduced to address the gap in video CMI effectiveness, utilizing a CNN for analyzing temporally distributed patches from video frames and employing a fusion system to consolidate forensic information. An ensemble classifier was introduced in \cite{wang2019ensemble} for source camera identification, leveraging fusion features to detect software-related, hardware-related, and statistical characteristics imprinted on images by digital cameras. In \cite{dal2021cnn}, an approach to CMI for video sequences was introduced, employing fusion techniques that leverage both audio and visual information within a multi-modal framework, demonstrating better performance over traditional mono-modal methods in tests conducted on the VISION dataset described in Section~\ref{se:dataset}.

\section{Dataset Description and Preparation}\label{se:dataset}

Here, the publicly available VISION dataset \cite{shullani2017vision} \cite{VISION} is utilized, comprising images and videos captured across various scenes and imaging conditions. As can be observed in Table~\ref{tab:vision}, a total of 35 camera devices, representing 29 camera models and 11 camera brands, are encompassed within this dataset. Specifically, there are 6 camera models featuring multiple instances per model, facilitating an investigation into the performance of the proposed approach at the device level.

VISION includes 648 native videos, which remain unaltered post-capture by the camera. These native videos were disseminated via social media platforms like YouTube and WhatsApp, with corresponding versions available in the dataset. Of the 684 native videos, 644 were shared via YouTube and 622 via WhatsApp. Upon being uploaded to YouTube, videos are compressed yet retain their initial resolutions, which span from $640 \times 480$ pixels for standard definition to as high as $1920 \times 1080$ pixels. In contrast, an alteration is observed when videos are shared on WhatsApp. Regardless of their original quality, they are rescaled to a resolution of $480 \times 848$ pixels. Through this process, the original video quality is often compromised on WhatsApp videos to ensure swift sharing and reduced data usage.

Moreover, the videos obtained from each camera are classified into three distinct scenarios: flat, indoor, and outdoor. Flat videos depict scenes with relatively homogeneous content, such as skies and white walls. Indoor scenarios encompass videos captured within indoor settings, such as offices and homes. Conversely, outdoor scenarios feature videos of gardens and streets. This diversity in scene content underscores the suitability of the VISION dataset as a benchmark for assessing source camera identification.

Taking into account the VISION dataset naming conventions outlined in \cite{shullani2017vision}, videos captured by devices D04, D12, D17, and D22 are excluded due to issues encountered during frame extraction or audio track retrieval. 

The VISION dataset is partitioned into training, testing, and validation sets to conduct a typical five-fold stratified cross-validation so that the standard deviation of accuracy is estimated. The choice of 5 folds is a compromise between an acceptable estimation of the standard deviation of accuracy and computational time. The standard deviation is reduced after fusion. This demonstrates the precision of the method. 

\section{Framework}\label{se:framework}

\subsection{Audio and Visual Content Feature Extraction}\label{subse:featureExtraction}

Our approach integrates audio and visual content to classify the videos within the VISION dataset. A description of the features extracted from the audio and visual content follows.

\textbf{Audio content.} This phase encompasses extracting audio data from each video sequence and the computation of the log-Mel spectrogram. The log-Mel representation of each extracted audio is computed using three distinct windows and hop sizes. This results in a 3-channel log-Mel spectrogram that captures various frequency details, serving as a comprehensive feature representation for the CMI task. 

The log-Mel spectrograms are computed as follows. The Short-Time Fourier Transform (STFT) is performed on the audio signal, segmenting it into overlapping frames and providing a spectrogram representation of the signal's frequency content over time. Mathematically, the STFT of the input signal $x[n]$ is expressed as

\begin{equation}
X(m, f)=\sum_{n=-\infty}^{\infty} x[n] \; w[n-m] \; e^{-j 2 \pi f n},
\end{equation}
where $X(m,f)$ denotes the STFT at a specific time index $m$ and frequency $f$, with $w[n-m]$ representing the window function applied to the signal. The outcome of the STFT is a two-dimensional representation of the signal $x[n]$, $\pmb{X}$ of size $T \times F$, with $T$ denoting the number of temporal samples (i.e., overlapping frames) and $F$ standing for the number of frequency bins. $\pmb{X}$ is referred to as the spectrogram of signal $x[n]$, having as elements the magnitude of the STFT.   

Following the STFT, the frequencies are transformed onto the Mel scale to produce the Mel spectrogram. This involves converting linear frequencies to the Mel scale using the expression
\begin{equation}
\operatorname{Mel}(f)=2595 \cdot \log _{10}\left(1+\frac{f}{700}\right).
\end{equation}
Then, a series of triangular filters based on these Mel frequencies are applied to the magnitude spectrum of the STFT. 
The Mel filter bank is denoted by a two-dimensional matrix $\pmb{H}$ of size $F \times K$, where $K$ is the number of triangular filters.
The triangular Mel filters, each centered at a Mel frequency corresponding to a pitch $p$, are defined as

\begin{equation}
\pmb{H}_p(f) = \begin{cases}
\frac{f-f_{p-1}}{f_{p}-f_{p-1}} & \text{for } f_{p-1} \leq f < f_p \\
\frac{f_{p+1}-f}{f_{p+1}-f_p} & \text{for } f_p \leq f < f_{p+1} \\
0 & \text{otherwise},
\end{cases}
\end{equation}
where $f_{p}=\operatorname{Mel}^{-1}(p)$ represents the center frequency of the filter corresponding to pitch $p$, and $f_{p-1}$ and $f_{p+1}$ are the center frequencies of the immediately adjacent filters.

Finally, the Mel spectrogram is converted into a log-Mel spectrogram by applying a logarithmic transformation to its values
\begin{equation}
\text{Log-Mel Spectrogram} = \pmb{L}= \ln( \pmb{X}  \pmb{H} + \epsilon),
\end{equation}
where $\epsilon$ is a small constant added to prevent zero values. This logarithmic transformation mirrors the logarithmic nature of human loudness perception, ensuring that the resulting log-Mel spectrogram closely aligns with human auditory processing. 

\textbf{Visual content.} This stage involves extracting video frames and preprocessing them by resizing them to a predefined size of $256 \times 256 \times 3$. Here, we use the raw video frames without performing any feature extraction, such as PRNU analysis.

\begin{figure*}[!ht]
    \centering
    \includegraphics[width=0.6\textwidth]{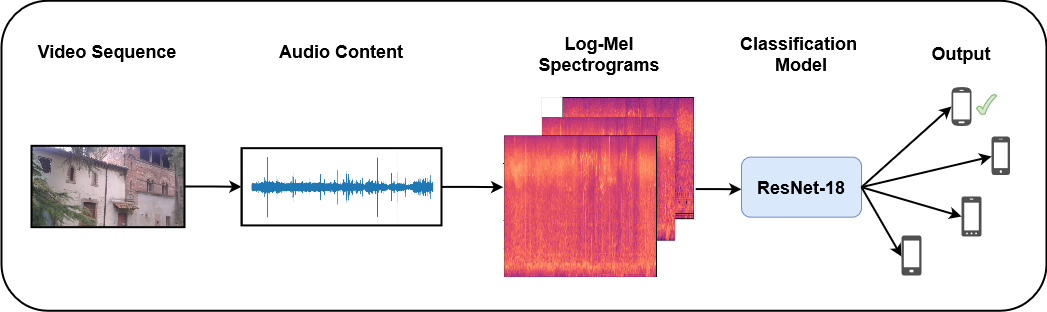}
    \caption{Flowchart depicting the CMI using only the audio content.}
    \label{fig:audio_flowchart}
\end{figure*}

\begin{figure*}[!ht]
    \centering
    \includegraphics[width=0.6\textwidth]{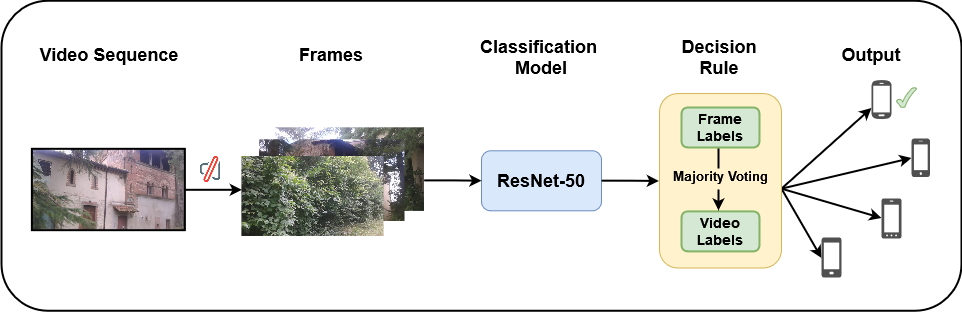}
    \caption{Flowchart depicting the CMI employing frames extracted from video sequences.}
    \label{fig:video_flowchart}
\end{figure*} 

\subsection{Unimodal Classification Methodology}\label{subse:unimodal}

Let us consider a scenario where a pattern needs to be assigned to one of the classes $\{\mathcal{C}_c\}_{c=1}^{C}$. Furthermore, let $\{\pmb{\gamma}_m\}_{m=1}^M$ be the set of random variables whose instances represent data samples of the $m$th modality. We denote the instances of the $m$th modality as 
$\{\pmb{\gamma}_m^{(n)}\}_{n=1}^{N}$. 

Furthermore, let $\circ$ be the function composition. If the classification system of the $m$th modality is realized by a neural network of $L$ layers, we can denote its output activation as
\begin{equation}
\pmb{a}_m^{(n)[L]} = \left(f_{\pmb{W}_m^{[L]}}^{[L]}\circ f_{\pmb{W}_m^{[L-1]}}^{[L-1]} \circ \cdots f_{\pmb{W}_m^{[1]}}^{[1]}\right) (\pmb{\gamma}_m^{(n)}),
\end{equation} 
where $\pmb{W}_m^{[l]}$ and $f_{\pmb{W}_m^{[l]}}^{[l]}$ are the parameters and the activation function of the $l$th layer, respectively.

Consider the collection of parameters belonging to the $L$th layer where each element is associated with the $c'$th classification node $\{\pmb{w}_m^{c',[L]}\}_{c'=1}^C$.  Also, let $\exp(\cdot)$ be the exponential function.
When the output activation function $f^{[L]}$ is the softmax function, the classification probabilities of the $c'$ classification node are given by
\begin{equation}\label{eq:softmaxProb}
    \Pr(\mathcal{C}_{c'} \mid \pmb{\gamma}_m^{(n)}\mathrel{;}\pmb{w}_m^{c',[L]}) = \frac{\exp \left({\pmb{w}_m^{c',[L]}}^\top \pmb{a}_m^{(n)[L-1]}\right)}{\sum_{c=1}^C \exp \left( {\pmb{w}_m^{c,[L]}}^\top \pmb{a}_m^{(n)[L-1]}\right) }.
\end{equation} In addition, the classification probabilities of the $n$th sample $\pmb{\gamma}_m^{(n)}$ related to the $m$th modality are given by
\begin{equation}\label{eq:hm}
    \pmb{p}_m^{(n)[L]} = \begin{bmatrix}
        \Pr(\mathcal{C}_1 \mid \pmb{\gamma}_m^{(n)} \mathrel{;} \pmb{w}_m^{1, [L]}) \\
        \Pr(\mathcal{C}_2 \mid \pmb{\gamma}_m^{(n)} \mathrel{;} \pmb{w}_m^{2, [L]}) \\ 
        \vdots \\
        \Pr(\mathcal{C}_C \mid \pmb{\gamma}_m^{(n)} \mathrel{;} \pmb{w}_m^{C, [L]})
    \end{bmatrix} \in \mathbb{R}^C.
\end{equation} In the remaining analysis, for simplicity, the superscript ${[L]}$ is omitted. Given the samples $\{\pmb{\gamma}_m^{(n)}\}_{n=1}^{N}$ of the $m$th modality, we obtain
\begin{equation}\label{eq:modalityProbs}
    \pmb{P}_m = [\pmb{p}_m^{(1)}, \pmb{p}_m^{(2)}, \dots, \pmb{p}_m^{(N)}]\in \mathbb{R}^{C \times N}.
\end{equation}

\noindent \textbf{Loss function.} Let $\pmb{T} = [\pmb{t}^{(1)}, \dots, \pmb{t}^{(N)}]\in \mathbb{R}^{C \times N}$ be the matrix of target variables. The $(c', n)$ element of $\pmb{T}$ is denoted by $t^{(n)}_{c'}$. The target vector $\pmb{t}^{(n)}$, that corresponds to the sample $\pmb{\gamma}_m^{(n)}$, adheres to the one-hot encoding scheme. In this scheme, if $\pmb{\gamma}_m^{(n)}$ belongs to class $\mathcal{C}_{c'}$, the target vector $\pmb{t}^{(n)}$ has zero elements except for the $c'$th element, which is set to one. In the proposed framework, the cross-entropy loss 
\begin{equation}
E \left(\{\pmb{\gamma}^{(n)}\}_{n=1}^{N}, \{\pmb{W}_m^{[l]}\}_{l=1}^L \right) = -\sum_{n=1}^N \sum_{c'=1}^C t_{c'}^{(n)}\ln \: [\pmb{p}_m^{(n)}]_{c'},
\end{equation} is used by the $m$th classification system.

\noindent \textbf{Unimodal Training} Our objective is to identify the camera model of each video within the VISION dataset. This task is treated as a classification problem, where each class in $\{\mathcal{C}_{c=1}^{C}\}$ refers to the IDs in Table~\ref{tab:vision}, with $C=25$. Each video is characterized by a single audio file and multiple video frames prepossessed following the guidelines in Section~\ref{subse:featureExtraction}. The audio files are related to the audio content modality ($m=1$), while the video frames are associated with the visual content modality ($m=2$). Given this distinction, two separate CNNs are trained, one for each modality.

To classify the audio files into one of the classes $\{\mathcal{C}_{c=1}^{25}\}$, a \texttt{ResNet18} model~\cite{he2016deep} is utilized. The $n$th audio file, denoted by $\pmb{\gamma}_1^{(n)}$, is assigned a vector of classification probabilities represented by $\pmb{p}_1^{(n)}$. Figure~\ref{fig:audio_flowchart} depicts the flow chart of the CMI using only the audio content. 

Similarly, to classify the video frames into one of the classes $\{\mathcal{C}_{c=1}^{25}\}$, a \texttt{ResNet50}~\cite{he2016deep} model is utilized. 
The $n$th video file, denoted by $\pmb{\gamma}_2^{(n)}$, is assigned a vector of classification probabilities represented by $\pmb{p}_2^{(n)}$. As the \texttt{ResNet50} model computes a probability vector for each video frame, $\pmb{p}_2^{(n)}$ is calculated as the average probability vector of all frames of the $n$th video. Figure~\ref{fig:video_flowchart} depicts the flow chart of the CMI using only the video content.

\noindent \textbf{Unimodal Testing Procedure.} The predicted classes of each sample $\{\pmb{\gamma}_m^{(n)}\}_{n=1}^{N}$ are given by
\begin{equation}\label{eq:predVector}
\pmb{c}_m = [\pazocal{C}_m^1, \pazocal{C}_m^2, \dots, \pazocal{C}_m^N]^\top \in \mathbb{R}^N,
\end{equation} where 
\begin{equation}\label{eq:argmaxProb}
    \pazocal{C}_{m}^n = \argmax_{c=1,\dots,C} \left[\pmb{p}_m^{(n)}\right]_{c},
\end{equation} is the predicted class of the $n$th sample with $\pazocal{C}_m^{n} \in \{\mathcal{C}_{c=1}^{C}\}$. 

\subsection{Multimodal Classification Methodology}\label{subse:fussion}

Multi-modal deep learning has demonstrated effectiveness in previous studies \cite{ngiam2011,liu2018}. Here, we utilize the product and sum rule for late fusion \cite{kittler1998}. Note that late fusion occurs subsequent to training classification models, which are utilized to generate classification probabilities for each sample. The \textit{product rule} is given by
\begin{equation}
    \pmb{P}_{\text{prod}} =  \pmb{P}_1 \odot \pmb{P}_2 \odot \cdots \odot \pmb{P}_M \in \mathbb{R}^{C \times N},
\end{equation} 
where $\odot$ denotes the Hadamard, element-wise, product. The \textit{sum rule} is given by
\begin{equation}
    \pmb{P}_{\text{sum}} = \pmb{P}_1 + \pmb{P}_2 + \cdots + \pmb{P}_M \in \mathbb{R}^{C \times N}.
\end{equation} 
\noindent \textbf{Testing Procedure.} After performing late fusion, the predicted class for each sample in $\{ \pmb{\gamma}_m^{(n)}\}_{n=1}^{N}$ is determined by applying (\ref{eq:argmaxProb}) to $\pmb{P}_{\text{prod}}$ or $\pmb{P}_{\text{sum}}$. This process yields the classification results obtained using the product or sum rule, respectively.

\section{Experimental Evaluation}\label{se:evaluation}

Table~\ref{tab:AudioVisionUnimodal} summarizes the results when the visual and the audio content are used separately. As can be seen, the mean accuracy using visual content in the Native, WhatsUp, and YouTube is $88.24\%$, $69.43\%$, and $71.77\%$, respectively. When audio content is used, the mean accuracy in the Native, WhatsUp, and YouTube is $93.99\%$, $91.11\%$, and $91.89\%$, respectively.

\captionsetup{font={footnotesize,sc},justification=centering,labelsep=period}%
\begin{table}[!ht]
    \centering
    \caption{Accuracy (\%) results using visual and audio content}
    \label{tab:AudioVisionUnimodal}
    \resizebox{\columnwidth}{!}{%
    \begin{tabular}{c*{6}{c}}
        \toprule
        & \multicolumn{3}{c}{\textbf{Visual-ResNet-50}} & \multicolumn{3}{c}{\textbf{Audio-ResNet-18}} \\
        \cmidrule(lr){2-4} \cmidrule(lr){5-7}
        & Native & WhatsApp & YouTube & Native & WhatsApp & YouTube \\
        \midrule 
        Fold 0  & 88.31  & 67.53 & 77.02 & 96.10 & 93.50 & 91.9  \\
        Fold 1  & 85.70  & 83.11 & 72.97 & 94.80 & 90.90 & 93.24 \\
        Fold 2  & 89.60  & 63.63 & 77.02 & 90.90 & 88.31 & 95.94 \\
        Fold 3  & 89.47  & 68.42 & 63.51 & 93.42 & 94.73 & 82.43 \\
        Fold 4  & 88.15  & 64.47 & 78.37 & 94.73 & 88.15 & 95.94 \\ \midrule
\makecell{Mean \\ $\pm$ StD}  & \makecell{88.24 \\ $\pm$ 1.4} & \makecell{69.43 \\ $\pm$ 7.07} & \makecell{71.77 \\ $\pm$ 5.44} & \makecell{93.99 \\ $\pm$ 1.76} & \makecell{91.11 \\ $\pm$ 2.66} & \makecell{91.89 \\ $\pm$ 4.98} \\
        \bottomrule
    \end{tabular}
    }
\end{table}
\captionsetup{font={footnotesize,rm},justification=centering,labelsep=period}%

Table~\ref{tab:AudioVisionFussion} summarizes the results achieved by applying late fusion on the outcomes obtained by the classifiers related to the visual and audio content. The late fusion uses the product or sum rule described in Section~\ref{subse:fussion}. As can be seen, the mean accuracy using the product rule in the Native, WhatsUp, and YouTube is $97.64\%$, $92.93\%$, and $95.59\%$, respectively. When the sum rule is used, the mean accuracy in the Native, WhatsUp, and YouTube is $96.33\%$, $93.72\%$, and $93.77\%$, respectively.

\captionsetup{font={footnotesize,sc},justification=centering,labelsep=period}%
\begin{table}[!ht]
    \centering
    \caption{Accuracy (\%) results using the product and sum rule}
    \label{tab:AudioVisionFussion}
    \resizebox{\columnwidth}{!}{%
    \begin{tabular}{c*{6}{c}}
        \toprule
        & \multicolumn{3}{c}{\textbf{Product Rule}} & \multicolumn{3}{c}{\textbf{Sum Rule}}\\
        \cmidrule(lr){2-4} \cmidrule(lr){5-7}
        & Native & WhatsApp & YouTube & Native & WhatsApp & YouTube \\
        \midrule 
        Fold 0  & 97.40 & 94.80 & 95.94 & 97.40 & 96.10 & 94.59 \\
        Fold 1  & 97.40 & 94.80 & 94.59 & 96.10 & 96.10 & 93.24 \\
        Fold 2  & 98.70 & 93.50 & 95.94 & 97.40 & 90.90 & 97.29 \\
        Fold 3  & 97.36 & 94.73 & 90.54 & 94.73 & 97.36 & 86.48 \\
        Fold 4  & 97.36 & 86.84 & 95.94 & 96.05 & 88.15 & 97.29 \\ \midrule
\makecell{Mean \\ $\pm$ StD} & \makecell{97.64 \\ $\pm$ 0.52} & \makecell{92.93 \\$\pm$ 3.08} & \makecell{95.59 \\ $\pm$ 0.52} & \makecell{96.33 \\ $\pm$ 0.99} & \makecell{93.72\\ $\pm$ 3.56} & \makecell{93.77 \\ $\pm$ 3.97} \\
        \bottomrule
    \end{tabular}
    }
\end{table}
\captionsetup{font={footnotesize,rm},justification=centering,labelsep=period}%

Comparing the results in Tables~\ref{tab:AudioVisionUnimodal} and~\ref{tab:AudioVisionFussion}, when the product rule performs the fusion, the mean accuracy in the Native, WhatsUp, and Youtube is improved by $9.4\%$, $23.5\%$, and $23.82\%$, respectively. When the sum rule performs the fusion, the accuracy results in the Native, WhatsUp, and YouTube are improved by $2.34\%$, $2.61\%$, and $1.88\%$, respectively. In summary, combining the classification probabilities obtained from visual and audio content demonstrates a promising improvement in classification performance.

\captionsetup{font={footnotesize,sc},justification=centering,labelsep=period}%
\begin{table}[!ht]
    \centering
    \caption{
    McNemar’s $p$-values to evaluate the null hypothesis $H_{0,1}$
    }    
    \label{tab:SumProductFusionPvalues}
    \begin{tabular}{c*{4}{c}}
        \toprule
        Folds & Native & WhatsApp & YouTube \\
        \midrule 
        Fold 0  & 0.0 & 1.0 & 1.0 \\
        Fold 1  & 1.0 & 1.0 & 1.0 \\
        Fold 2  & 1.0 & 0.5 &  1.0 \\
        Fold 3  & 0.5 & 0.5 & 0.3 \\
        Fold 4  & 1.0 & 1.0 & 1.0 \\ 
        \bottomrule
    \end{tabular}
\end{table}
\captionsetup{font={footnotesize,rm},justification=centering,labelsep=period}%

Next, we study the null hypotheses: 
\begin{itemize}
    \item $H_{0,1}$: The classification performances achieved by the two fusion rules are equivalent.
    \item $H_{0,2}$: The classification performance achieved solely with visual content is equivalent to that achieved with the product rule.
    \item $H_{0,3}$: The classification performance achieved solely with audio content is equivalent to that achieved with the product rule.
\end{itemize} We have significant evidence or highly significant evidence against $H_{0,i}$, for $i=1,2,3$, when the $p$-value falls within the range $[0.01, 0.05]$ or $[0, 0.01]$, respectively. When $p$-value is greater that $0.05$, we have not a significant evidence against $H_{0,i}$, for $i = 1, 2, 2$. Here, $p$-values are computed by applying McNemar's significance test~\cite{mcnemar1947} \cite{mcnemar_test}.

\captionsetup{font={footnotesize,sc},justification=centering,labelsep=period}%
\begin{table}[!ht]
    \centering
    \caption{
    McNemar’s $p$-values to evaluate the null hypotheses $H_{0,2}$ and $H_{0,3}$}
    \label{tab:AudioVisionFusionMcNemarPvalues}
    \resizebox{\columnwidth}{!}{%
    \begin{tabular}{c*{6}{c}}
        \toprule
        & \multicolumn{3}{c}{\textbf{Visual-ResNet-50}} & \multicolumn{3}{c}{\textbf{Audio-ResNet-18}} \\
        \cmidrule(lr){2-4} \cmidrule(lr){5-7}
        & Native & WhatsApp & YouTube & Native & WhatsApp & YouTube \\
        \midrule 
        Fold 0  & 0.023  & $10^{-5}$  & 0.001     & 1.0   & 1.0   & 0.371 \\
        Fold 1  & 0.007  & 0.026      & 0.001     & 0.617 & 0.248 & 1.0 \\
        Fold 2  & 0.044  & $10^{-5}$  & 0.001     & 0.041 & 0.133 & 0.479 \\
        Fold 3  & 0.041  & $10^{-5}$  & $10^{-5}$ & 0.248 & 0.617 & 0.007  \\
        Fold 4  & 0.045  & $10^{-4}$  & 0.002     & 0.617 & 1.0   & 0.479 \\
        \bottomrule
    \end{tabular}
}
\end{table}

Table~\ref{tab:SumProductFusionPvalues} summarizes the computed $p$-values for $H_{0,1}$. Most of the $p$-values exceed the predetermined significance threshold, so we lack significant evidence against $H_{0,1}$. Table~\ref{tab:AudioVisionFusionMcNemarPvalues} summarizes the computed $p$-values for $H_{0,2}$. It is evident that we have significant evidence against $H_{0,2}$. Table~\ref{tab:AudioVisionFusionMcNemarPvalues} summarizes also the computed $p$-values for $H_{0,3}$. Most of the $p$-values exceed the predetermined significance threshold, so we lack significant evidence against $H_{0,3}$.

\captionsetup{font={footnotesize,rm},justification=centering,labelsep=period}%

\section{Discussion and Future work}

Unlike~\cite{dal2021cnn} which analyzes smaller segments (patches) extracted from video frames and log-mel spectrograms, our framework utilizes the entirety of these data sources for prediction. While this difference in the prediction process prevents a direct comparison, we still report the accuracy results achieved by~\cite{dal2021cnn} to provide a general sense of our framework potential. 

The proposed framework achieves a mean accuracy of $76.31\%$ and $92.33\%$ when the visual and audio content is used, respectively, in Table~\ref{tab:AudioVisionUnimodal}. The mean accuracy is computed across the categories Native, WhatsApp, and YouTube. The corresponding accuracies in~\cite{dal2021cnn} for the visual and audio content are $74.84\%$ and $67.81\%$, respectively.

Regarding the fusion results returned by the proposed framework, the best mean accuracy across the Native, WhatsApp, and YouTube categories in Table~\ref{tab:AudioVisionFussion} is $95,38\%$. The latter accuracy is achieved by the product rule. The corresponding accuracy in~\cite{dal2021cnn} is $95.27\%$. 

Both unimodal and bimodal classification indicate the potential of our approach for CMI, with the product rule demonstrating better performance than the sum rule. The superior performance of the product rule can be attributed to the higher joint probabilities generated when all modalities align, as observed in the mean results presented in Table~\ref{tab:AudioVisionUnimodal}.

Future work will focus on various key areas to further analyse our framework. The robustness of the framework can be investigated on different levels of noise. Possible overfitting issues can be analyzed by performing training with more lightweight models~\cite{wang2022lightweight}. Other datasets that contain more recent devices, like the FloreView dataset~\cite{baracchi2023}, can be employed to evaluate the proposed framework. 

\section{Conclusion}\label{se:conclusion}

CMI holds significant importance in multimedia forensic applications. This paper introduces a framework capable of device identification using audio, visual content, or a combination of both. CNNs are employed to address the device identification problem as a classification task. Experimental evaluation demonstrates a promising classification accuracy when independently using audio or visual content. Additionally, combining audio and visual content may lead to notable enhancements in classification performance, suggesting a potential area for further research. 

\section{Acknowledgments}
This research was supported by the Hellenic Foundation for Research and Innovation (HFRI) under the ``2nd Call for HFRI Research Projects to support Faculty Members \& Researchers" (Project Number: 3888).

\bibliographystyle{IEEEtran}
\bibliography{references}

\begin{thebibliography}{10}
\providecommand{\url}[1]{#1}
\csname url@samestyle\endcsname
\providecommand{\newblock}{\relax}
\providecommand{\bibinfo}[2]{#2}
\providecommand{\BIBentrySTDinterwordspacing}{\spaceskip=0pt\relax}
\providecommand{\BIBentryALTinterwordstretchfactor}{4}
\providecommand{\BIBentryALTinterwordspacing}{\spaceskip=\fontdimen2\font plus
\BIBentryALTinterwordstretchfactor\fontdimen3\font minus \fontdimen4\font\relax}
\providecommand{\BIBforeignlanguage}[2]{{%
\expandafter\ifx\csname l@#1\endcsname\relax
\typeout{** WARNING: IEEEtran.bst: No hyphenation pattern has been}%
\typeout{** loaded for the language `#1'. Using the pattern for}%
\typeout{** the default language instead.}%
\else
\language=\csname l@#1\endcsname
\fi
#2}}
\providecommand{\BIBdecl}{\relax}
\BIBdecl

\bibitem{berdich2023survey}
A.~Berdich, B.~Groza, and R.~Mayrhofer, ``A survey on fingerprinting technologies for smartphones based on embedded transducers,'' \emph{IEEE Internet of Things Journal}, vol.~10, no.~16, pp. 14\,646--14\,670, 2023.

\bibitem{nwokeji2024source}
C.~E. Nwokeji, A.~Sheikh-Akbari, A.~Gorbenko, and I.~Mporas, ``Source camera identification techniques: A survey,'' \emph{Journal of Imaging}, vol.~10, no.~2, p.~31, 2024.

\bibitem{stamm2013information}
M.~C. Stamm, M.~Wu, and K.~J.~R. Liu, ``Information forensics: An overview of the first decade,'' \emph{IEEE Access}, vol.~1, pp. 167--200, 2013.

\bibitem{diwan2023visualizing}
A.~Diwan and U.~Sonkar, ``Visualizing the truth: A survey of multimedia forensic analysis,'' \emph{Multimedia Tools and Applications}, pp. 1--28, 2023.

\bibitem{lukas2006digital}
J.~Lukas, J.~Fridrich, and M.~Goljan, ``Digital camera identification from sensor pattern noise,'' \emph{IEEE Transactions on Information Forensics and Security}, vol.~1, no.~2, pp. 205--214, 2006.

\bibitem{davis1980comparison}
S.~Davis and P.~Mermelstein, ``Comparison of parametric representations for monosyllabic word recognition in continuously spoken sentences,'' \emph{IEEE Transactions on Acoustics, Speech, and Signal Processing}, vol.~28, no.~4, pp. 357--366, 1980.

\bibitem{kotropoulos2014source}
C.~Kotropoulos, ``Source phone identification using sketches of features,'' \emph{IET biometrics}, vol.~3, no.~2, pp. 75--83, 2014.

\bibitem{kotropoulos2014mobile}
C.~Kotropoulos and S.~Samaras, ``Mobile phone identification using recorded speech signals,'' in \emph{Proceedings of the 19th International Conference on Digital Signal Processing}.\hskip 1em plus 0.5em minus 0.4em\relax IEEE, 2014, pp. 586--591.

\bibitem{Kritsiolis2024}
D.~Kritsiolis and C.~Kotropoulos, ``Mobile phone identification from recorded speech signals using non-speech segments and universal background model adaptation,'' in \emph{Proceedings of the 13th International Conference on Pattern Recognition Applications and Methods}, 2024, pp. 793--800.

\bibitem{bondi2016first}
L.~Bondi \emph{et~al.}, ``First steps toward camera model identification with convolutional neural networks,'' \emph{IEEE Signal Processing Letters}, vol.~24, no.~3, pp. 259--263, 2016.

\bibitem{kittler1998}
J.~Kittler, M.~Hatef, R.~P. Duin, and J.~Matas, ``On combining classifiers,'' \emph{IEEE Transactions on Pattern Analysis and Machine Intelligence}, vol.~20, no.~3, pp. 226--239, 1998.

\bibitem{IARIADevIDFusion}
\BIBentryALTinterwordspacing
``Camera model identification fusing audio and visual content,'' [retrieved: May 20, 2024]. [Online]. Available: \url{https://github.com/iTsingalis/IARIADevIDFusion}
\BIBentrySTDinterwordspacing

\bibitem{timmerman2020video}
D.~Timmerman, S.~Bennabhaktula, E.~Alegre, and G.~Azzopardi, ``Video camera identification from sensor pattern noise with a constrained convnet,'' \emph{arXiv preprint arXiv:2012.06277}, 2020.

\bibitem{lopez2017smartphone}
R.~R. L{\'o}pez, A.~El-Khattabi, A.~L.~S. Orozco, and L.~J.~G. Villalba, ``Smartphone video source identification based on sensor pattern noise,'' \emph{International Journal of Electronics and Communication Engineering}, vol.~11, no.~5, pp. 597--600, 2017.

\bibitem{mandelli2020modified}
S.~Mandelli \emph{et~al.}, ``A modified {Fourier-Mellin} approach for source device identification on stabilized videos,'' in \emph{Proceedings of the International Conference on Image Processing}.\hskip 1em plus 0.5em minus 0.4em\relax IEEE, 2020, pp. 1266--1270.

\bibitem{altinisik2022video}
E.~Altinisik, H.~T. Sencar, and D.~Tabaa, ``Video source characterization using encoding and encapsulation characteristics,'' \emph{IEEE Transactions on Information Forensics and Security}, vol.~17, pp. 3211--3224, 2022.

\bibitem{akbari2022prnu}
Y.~Akbari, N.~Almaadeed, S.~Al-Maadeed, F.~Khelifi, and A.~Bouridane, ``{PRNU-net}: A deep learning approach for source camera model identification based on videos taken with smartphone,'' in \emph{Proceedings of the 26th International Conference on Pattern Recognition}.\hskip 1em plus 0.5em minus 0.4em\relax IEEE, 2022, pp. 599--605.

\bibitem{bennabhaktula2022source}
G.~S. Bennabhaktula, D.~Timmerman, E.~Alegre, and G.~Azzopardi, ``Source camera device identification from videos,'' \emph{SN Computer Science}, vol.~3, no.~4, p. 316, 2022.

\bibitem{shullani2017vision}
D.~Shullani, M.~Fontani, M.~Iuliani, O.~A. Shaya, and A.~Piva, ``Vision: A video and image dataset for source identification,'' \emph{EURASIP Journal on Information Security}, vol. 2017, pp. 1--16, 2017.

\bibitem{akbari2022new}
Y.~Akbari \emph{et~al.}, ``A new forensic video database for source smartphone identification: Description and analysis,'' \emph{IEEE Access}, vol.~10, pp. 20\,080--20\,091, 2022.

\bibitem{manisha2023source}
N.~Manisha, C.-T. Li, and K.~A. Kotegar, ``Source camera identification with a robust device fingerprint: Evolution from image-based to video-based approaches,'' \emph{Sensors}, vol.~23, no.~17, p. 7385, 2023.

\bibitem{hosler2019video}
B.~Hosler \emph{et~al.}, ``A video camera model identification system using deep learning and fusion,'' in \emph{Proceedings of the IEEE International Conference on Acoustics, Speech, and Signal Processing}.\hskip 1em plus 0.5em minus 0.4em\relax IEEE, 2019, pp. 8271--8275.

\bibitem{wang2019ensemble}
B.~Wang, K.~Zhong, and M.~Li, ``Ensemble classifier-based source camera identification using fusion features,'' \emph{Multimedia Tools and Applications}, vol.~78, no.~7, pp. 8397--8422, 2019.

\bibitem{dal2021cnn}
D.~Dal~Cortivo, S.~Mandelli, P.~Bestagini, and S.~Tubaro, ``{CNN}-based multi-modal camera model identification on video sequences,'' \emph{Journal of Imaging}, vol.~7, no.~8, p. 135, 2021.

\bibitem{VISION}
\BIBentryALTinterwordspacing
``Vision dataset,'' [retrieved: May 20, 2024]. [Online]. Available: \url{https://lesc.dinfo.unifi.it/VISION/}
\BIBentrySTDinterwordspacing

\bibitem{he2016deep}
K.~He, X.~Zhang, S.~Ren, and J.~Sun, ``Deep residual learning for image recognition,'' in \emph{Proceedings of the IEEE Conference on Computer Vision and Pattern Recognition}, 2016, pp. 770--778.

\bibitem{ngiam2011}
J.~Ngiam \emph{et~al.}, ``Multimodal deep learning,'' in \emph{Proceedings of the International Conference on Machine Learning}, 2011, pp. 689--696.

\bibitem{liu2018}
K.~Liu, Y.~Li, N.~Xu, and P.~Natarajan, ``Learn to combine modalities in multimodal deep learning,'' \emph{arXiv preprint arXiv:1805.11730}, 2018.

\bibitem{mcnemar1947}
Q.~{McNemar}, ``Note on the sampling error of the difference between correlated proportions or percentages,'' \emph{Psychometrika}, vol.~12, no.~2, pp. 153--157, 1947.

\bibitem{mcnemar_test}
\BIBentryALTinterwordspacing
``{McNemar's} test for classifier comparisons,'' [retrieved: May 20, 2024]. [Online]. Available: \url{https://rasbt.github.io/mlxtend/user_guide/evaluate/mcnemar/}
\BIBentrySTDinterwordspacing

\bibitem{wang2022lightweight}
C.-H. Wang, K.-Y. Huang, Y.~Yao, J.-C. Chen, H.-H. Shuai, and W.-H. Cheng, ``{Lightweight Deep Learning: An Overview},'' \emph{IEEE Consumer Electronics Magazine}, 2022.

\bibitem{baracchi2023}
D.~Baracchi, D.~Shullani, M.~Iuliani, and A.~Piva, ``{FloreView}: {A}n image and video dataset for forensic analysis,'' \emph{IEEE Access}, 2023.

\end{thebibliography}

\end{document}